\documentclass[pdflatex,sn-basic]{sn-jnl}
\usepackage{graphicx}%
\usepackage{multirow}%
\usepackage{amsmath,amssymb,amsfonts}%
\usepackage{amsthm}%
\usepackage[title]{appendix}%
\usepackage{xcolor}%
\usepackage{textcomp}%
\usepackage{manyfoot}%
\usepackage{booktabs}%
\usepackage{algorithm}%
\usepackage{algorithmicx}%
\usepackage{algpseudocode}%
\usepackage{listings}%
\usepackage[T1]{fontenc} % Font warnings 
\usepackage{bookmark} % hyperref warnings
\usepackage{makecell} % thead in tables
\usepackage{tabularx} % Appendix tables
\begin{document}
\title[MisSynth]{MisSynth: Improving MISSCI Logical Fallacies Classification with Synthetic Data}

\author{\fnm{Mykhailo} \sur{Poliakov}}\email{mykhailo.poliakov@ukma.edu.ua}
\author{\fnm{Nadiya} \sur{Shvai}}\email{n.shvay@ukma.edu.ua}
\affil{\orgname{National University of Kyiv-Mohyla Academy}, \orgaddress{\city{Kyiv}, \country{Ukraine}}}

\abstract{Health-related misinformation is very prevalent and potentially harmful. It is difficult to identify, especially when claims distort or misinterpret scientific findings. We investigate the impact of synthetic data generation and lightweight fine-tuning techniques on the ability of large language models (LLMs) to recognize fallacious arguments using the \textit{MISSCI} dataset and framework. In this work, we propose \textit{MisSynth}, a pipeline that applies retrieval-augmented generation (RAG) to produce synthetic fallacy samples, which are then used to fine-tune an LLM model. Our results show substantial accuracy gains with fine-tuned models compared to vanilla baselines. For instance, the LLaMA 3.1 8B fine-tuned model achieved an over 35\% F1-score absolute improvement on the \textit{MISSCI} test split over its vanilla baseline. We demonstrate that introducing synthetic fallacy data to augment limited annotated resources can significantly enhance zero-shot LLM classification performance on real-world scientific misinformation tasks, even with limited computational resources. The code and synthetic dataset are available on \href{https://github.com/mxpoliakov/MisSynth}{GitHub}.
}


\maketitle
\section{Introduction}
Health-related misinformation has been identified as one of the major factors that deteriorate global health and lead to a decrease in public trust in science \citep{brennen:2020}. This threat is growing because all forms of falsehood are spreading farther and faster than truth online \citep{vosoughi:2018}. The problem is especially dangerous when real scientific findings are distorted. For instance, misleading reports often use selective and deceptive quotations of scientific work to support false claims \citep{adversarial:2024}. On the other hand, discredited and retracted research continues to be mentioned as valid, supporting arguments with empty research \citep{zombie:2024}. These arguments use the credibility of the source to hide subtle logical fallacies.

Detecting such fallacies is a major challenge. It requires a deep understanding of scientific context and logical reasoning. Often, the flawed thinking shortcuts that make readers susceptible to fallacies are more intuitive than the deliberate analysis required to debunk them \citep{lewandowsky:2020}. Even the largest large language models (LLMs) can perform poorly on this task. One recent benchmark highlights this performance gap by testing for implicit fallacious reasoning \citep{missci:2024}. Other new datasets tools also show that LLMs lag far behind humans in identifying fine-grained fallacies \citep{hong:2024}. A comprehensive benchmark that unifies previous datasets further confirms these limitations \citep{mafalda:2024}. This performance gap can be attributed to the scarcity of large, high-quality annotated datasets for training.

We find that current methods are often insufficient. Traditional fact-checking systems are designed to find explicit counter-evidence \citep{nakov:2021}. Such systems are not suited for complex cases where evidence is slightly distorted rather than outright fabricated \citep{guo:2022}. Synthetic data can help address data scarcity \citep{moller:2024}. However, synthetic data often produces templated and unnatural examples. This creates a critical distribution gap that risks training models to excel at detecting AI-generated misinformation while leaving them vulnerable to the diverse and unpredictable real-world misinformation \citep{li:2024}.

We introduce \textit{MisSynth}, a new pipeline aiming to address this issue. Our novel technique employs retrieval-augmented generation (RAG) to produce realistic and context-sensitive synthetic data \citep{rag:2020}. We use this data to fine-tune an LLM with a parameter-efficient technique called Low-Rank Adaptation (LoRA) by \cite{lora:2021}. Our experiments show this approach yields significant gains. For example, a fine-tuned LLaMA 3.1 8B model improved its F1-score by over 35\% (absolute gain) on the \textit{MISSCI} test split. This demonstrates the effectiveness of our method, even with limited computational resources. Our primary contributions are as follows:
\begin{itemize}
    \item We present \textit{MisSynth}, a novel RAG-based pipeline for generating high-quality synthetic data of logical fallacies.
    \item We show that fine-tuning with our synthetic data significantly improves an LLM's performance on the logical fallacy classification subtask of the \textit{MISSCI} benchmark.
    \item We release the synthetic dataset generated by \textit{MisSynth} (GPT-5 version) publicly.
\end{itemize}

The main novelty is the integration of RAG with parameter-efficient fine-tuning (LoRA) specifically for logical fallacy classification. Unlike earlier data augmentation techniques that often produce templated or context-less examples, the \textit{MisSynth} pipeline enforces a same-source retrieval constraint. This crucial step ensures the generated synthetic arguments are grounded in the source scientific article and are realistic. By utilizing this pipeline, we introduce an efficient and effective method for specializing large language models for complex scientific reasoning tasks, particularly in scenarios where high-quality annotated data is scarce.

The rest of this paper is structured as follows. We first review related work. Then, we detail our methodology. Next, we present our experiments and results. Finally, we discuss our findings and suggest future research directions.
\section{Related Work}
\subsection{Fallacy Detection and Scientific Misinformation}
Detecting flawed reasoning in arguments is a significant challenge, particularly within the context of scientific misinformation \citep{wachsmuth:2017}. Traditional methods often fail due to scientific misinformation because fallacies are implicit and heavily dependent on context \citep{boudry:2015}. While recent work utilizes LLMs, their ability to classify subtle reasoning errors remains limited, underscoring the need for improved training data and methods \citep{ruiz-dolz:2023}. 
\subsection{Synthetic Data Generation}
Synthetic data generation offers a way to augment scarce training resources \citep{chung:2023,sennrich:2016}. However, some methods produce templated data, like the \textit{LFUD} dataset, which lacks the complexity of real-world arguments \citep{li:2024}. Our RAG-based approach generates more diverse and contextually grounded examples by drawing from authentic scientific texts.
\subsection{Fine-Tuning}
Full fine-tuning of large models is often impractical. Parameter-efficient fine-tuning (PEFT) methods provide an efficient alternative. We use Low-Rank Adaptation (LoRA), which freezes the pre-trained model and injects small, trainable matrices into its layers \citep{lora:2021}. This approach greatly reduces the number of trainable parameters and memory usage. LoRA allows effective fine-tuning on consumer-grade hardware without sacrificing performance.
\subsection{MISSCI}
Our work uses a recent benchmark designed for fallacy detection. The \textit{MISSCI} dataset provides a formal framework for our task \citep{missci:2024}. It models misinformation as an argument where an inaccurate claim, $\bar{c}$, is supported by an accurate premise, $P$, and a fallacious premise, $\bar{P}$. The accurate premise alone does not support the claim ($P \not\Rightarrow \bar{c}$), but the combination does ($P \cup \bar{P} \Rightarrow \bar{c}$). Each fallacious step is a triplet $R_{i}=(s_{i}, \bar{p}{i}, f{i})$, composed of scientific context $s_{i}$, a fallacious premise $\bar{p}{i}$, and a fallacy class $f{i}$. The complete argument is represented as:

\begin{equation}
\,\,  
\overset{\substack{s_0 \\ =}}{p_0}  
\,,\,  
\underset{\substack{\downarrow \\ f_1}}{\overset{\substack{s_1 \\ \downarrow}}{\overline{p}_1}}
\,,\,  
\ldots
\,,\,  
\underset{\substack{\downarrow \\ f_N}}{\overset{\substack{s_N \\ \downarrow}}{\overline{p}_N}}  
\,\,  
\,\,  
\Rightarrow \overline{c}
\end{equation}

The task involves identifying and classifying the fallacious premise $\bar{p}{i}$ and its type $f{i}$. Our work focuses on the classification part of \textit{MISSCI}. Its extension, \textit{MISSCIPlus}, incorporates additional arguments identified in the original scientific texts into the original dataset \citep{missciplus:2025}. Models must first find the relevant passage from the article before recognizing the fallacy. 
\subsection{Other Related Benchmarks}
Other notable benchmarks also confirm that LLMs struggle with nuanced argumentation, motivating our approach to improve model training. The \textit{LOGIC} dataset and its climate-focused subset, \textit{LogicClimate}, provide a general reasoning challenge for language models \citep{logic:2022}. The bilingual \textit{RuozhiBench} uses subtle logical inconsistencies to highlight the performance gap between LLMs and humans \citep{ruozhibench:2025}. The \textit{Fallacies} dataset offers a hierarchical taxonomy of over 200 fallacy types to assess the self-verification capabilities of LLMs \citep{hong:2024}. \textit{MAFALDA} unifies several previous datasets and introduces a "disjunctive annotation scheme" to account for the subjectivity of fallacy annotation by allowing multiple labels \citep{mafalda:2024}.

\section{Methodology}
\begin{figure}[ht!]
\includegraphics[width=\textwidth]{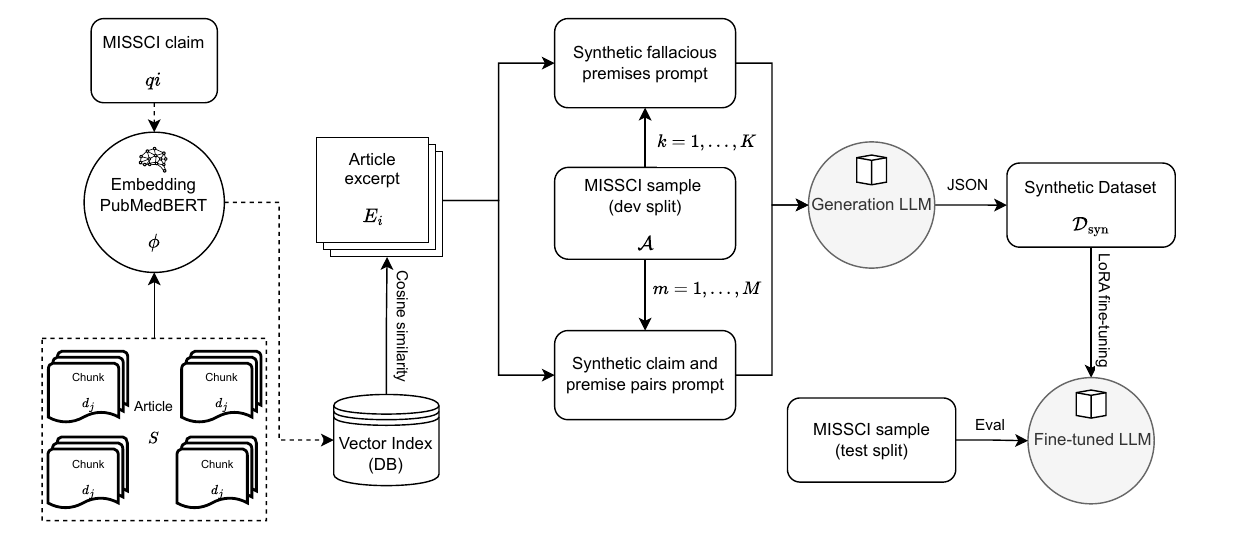}
 \caption{Overview of \textit{MisSynth} synthetic data generation and fine-tuning pipeline. A RAG retrieves an article excerpt ($E_i$) from a source article ($S$) based on a MISSCI claim ($q_i$). This excerpt, along with a dev split sample ($\mathcal{A}$), is used by a \textit{Generation LLM} to create a synthetic dataset ($\mathcal{D}_{\mathrm{syn}}$). This dataset is then used to fine-tune a model with LoRA, which is finally evaluated on the MISSCI test split.} \label{pipeline-figure}
\end{figure}

Detecting health misinformation that misuses scientific claims is a significant problem, partially due to the scarcity of real-world annotated data. This section describes \textit{MisSynth} methodology, which tackles the data shortage. We generate synthetic data to fine-tune Large Language Models (LLMs) for this task. The complete process is shown in Figure \ref{pipeline-figure}. Our method first retrieves relevant text using RAG (\ref{methodology-rag}). An LLM then uses this text to create new fallacy examples (\ref{methodology-synthetic-data}). We use this new dataset to locally fine-tune a model using LoRA (\ref{methodology-fine-tuning}). In addition, we detail our evaluation strategy (\ref{methodology-evaluation}). 
\subsection{RAG for publication context} \label{methodology-rag}
We base synthetic examples on the same publication contexts that produce fallacious reasoning in \textit{MISSCI}. For each instance in the dev split, we download the cited source $S$ and segment it with a recursive character splitter (chunk size 512, overlap 64). Each passage $d_j \in S$ is embedded with a PubMedBERT biomedical encoder by \cite{pubmedbert:2021} $\phi$, yielding

\begin{equation}
\mathbf{e}_j = \phi(d_j) \in \mathbb{R}^m
\end{equation}

We store the passages in the Langchain's \citep{langchain:2022} in-memory vector index along with metadata $\mathrm{source}(d_j) = u$. At retrieval time, we build the query from the inaccurate claim only,

\begin{equation}
q_i = \overline{c}, \qquad \mathbf{e}_{q_i} = \phi(q_i)
\end{equation}

and compute the cosine similarity:

\begin{equation}
\mathrm{sim}(q_i,d_j) = \frac{\mathbf{e}_{q_i}^\top \mathbf{e}_j}{\lVert \mathbf{e}_{q_i} \rVert \lVert \mathbf{e}_j \rVert}
\end{equation}

We then retrieve the top-$k$ passages subject to a same-source constraint, implemented as a metadata filter:

\begin{equation}
\mathcal{R}_k(q_i, u) = \operatorname*{arg\,topk}_{d_j \in \mathcal{D},\, \mathrm{source}(d_j) = u} \mathrm{sim}(q_i, d_j)
\end{equation}

with $k = 5$. We concatenate the retrieved passages into a single excerpt:

\begin{equation}
E_i = \mathrm{concat}\bigl(\mathcal{R}_k(q_i, u)\bigr).
\end{equation}

which is then used to generate the final answer. This setup follows retrieval with dual encoders (one for the query and one for the RAG passages) and top-$k$ similarity search, as well as a multi-passage sequence for generation \citep{rag:2020}. At the same time, the same-source filter enforces the \textit{MISSCI} assumption that fallacious reasoning is based on the cited source.

\subsection{Synthetic Data} \label{methodology-synthetic-data}

Let an annotated \textit{MISSCI} argument be $\mathcal{A} = (\overline{c}, p_0, R)$, where each reasoning step is $R_i = (s_i, \overline{p}_i, f_i)$ with publication context $s_i$, fallacious premise $\overline{p}_i$, and fallacy class $f_i$. For each dev instance, we extract the set of gold fallacies and their classes:

\begin{equation}
\mathcal{F}^{\mathrm{real}} = \left\{ (\overline{p}^{\mathrm{real}}_\ell, f^{\mathrm{real}}_\ell, s^{\mathrm{real}}_\ell) \right\}_\ell
\end{equation}

format a prompt with $(\overline{c}, p_0, \mathcal{F}^{\mathrm{real}}, E_i)$, and use a \textit{Generation LLM} to produce structured JSON. We generate two kinds of synthetic data:

\subsubsection{Synthetic fallacious premises.} We sample $K$ synthetic variants per dev split instance as triples of synthetic context, fallacious premise, and class:

\begin{equation}
(\tilde{s}_{i,k}, \tilde{\overline{p}}_{i,k}, \tilde{f}_{i,k}) \sim p_\theta\Bigl(s, \overline{p}, f \,\big|\, \overline{c}, p_0, \mathcal{F}^{\mathrm{real}}, E_i\Bigr), \qquad k = 1, \dots, K
\end{equation}

Each item must use a class from the fallacy inventory and be derived from the content of $E_i$. 

\subsubsection{Synthetic claim–premise pairs.} When enabled, we also sample $M$ coherent claim / accurate-premise pairs supported by the same source and excerpt:

\begin{equation}
(\tilde{c}_{i,m}, \tilde{p}_{0,i,m}) \sim p_\theta\bigl(c, p_0 \mid \mathcal{F}^{\mathrm{real}}, E_i\bigr), \qquad m = 1, \dots, M
\end{equation}

to increase diversity of inputs, since each $K$ fallacious premises from the above contain the same real claim–premise pairs per instance. 

\subsubsection{Prompting and parsing.} Prompts include the fallacy inventory (extracted from a template file) and require a strict JSON array with fields "context", "fallacy", and "class" (Appendices \ref{appendix-a}, \ref{appendix-b}). We skip instances with empty retrieval results or invalid JSON. The temperature of retrieval LLM is kept at $1.0$, where applicable.

\subsubsection{Train/validation set for fine-tuning.} We convert synthetic items into instruction–completion pairs using \textit{MISSCI}'s "classify with definition" template. For each synthetic fallacy:

\begin{equation}
x_{i,j} = T\bigl(\overline{c}, p_0, \tilde{s}_{i,j}, \tilde{\overline{p}}_{i,j}\bigr), \qquad
y_{i,j} = \text{"Fallacy: } \hat{f}_{i,j}\text{"}
\end{equation}

We form the training set $\mathcal{D}_{\mathrm{syn}} = \{(x_{i,j}, y_{i,j})\}$. The validation set uses only gold \textit{MISSCI} dev examples (original interchangeable fallacies) formatted with the same template, ensuring that validation contains no synthetic completions. 

We also include a \textit{random-baseline ablation} that replaces synthetic contexts and premises with lorem ipsum while keeping answers intact, to test whether gains come from synthetic content rather than prompt template or answer structure.

\subsection{Fine-tuning} \label{methodology-fine-tuning}
We adapt a frozen base model with parameters $\Phi_0$ using trainable LoRA \citep{lora:2021} parameters $\Theta$ while keeping $\Phi_0$ fixed. Let $\mathcal{Z}={(x,y)}$ denote the training pairs produced from $\mathcal{D}{\mathrm{syn}}$. We maximize the conditional likelihood with respect to the adapter parameters:
\begin{equation}
\max{\Theta} \sum_{(x,y)\in\mathcal{Z}} \sum_{t=1}^{|y|} \log\left( p_{\Phi_0+\Delta\Phi(\Theta)}\left(y_t \middle| x, y_{<t}\right)\right)
\end{equation}
In LoRA, the task-specific increment $\Delta\Phi(\Theta)$ is encoded by low-rank updates to selected linear projections, typically the attention projections $W_q$ and $W_v$. For any adapted weight $W_0 \in \mathbb{R}^{d\times k}$ in $\Phi_0$, we learn $A\in\mathbb{R}^{d\times r}$ and $B\in\mathbb{R}^{r\times k}$ with $r \ll \min{d,k}$ and set
\begin{equation}
W = W_0 + \frac{\alpha}{r} AB
\end{equation}
where only $A$ and $B$ are trainable and $\alpha$ is a fixed scaling factor. All other parameters of $\Phi_0$ remain frozen. 

\subsection{Evaluation} \label{methodology-evaluation}

We evaluate the \textit{MisSynth} pipeline by fine-tuning several LLMs using Low-Rank Adaptation (LoRA) and testing their performance on the classification sub-task of the \textit{MISSCI} benchmark. Our primary evaluation metrics are accuracy (Acc) and macro-averaged F1-score (F1) on the \textit{MISSCI} test split.

All fine-tuning experiments were performed using the LoRA technique with a rank of $r=8$ on the attention projections ($W_q$ and $W_v$). The training set $\mathcal{D}_{\mathrm{syn}}$ was generated from the \textit{MISSCI} dev set using the \textit{Generation LLM} at a temperature of 1.0, where applicable. We use the gold \textit{MISSCI} dev examples as a consistent fine-tuning validation set across all runs (96 samples). All experiments were conducted locally on an M1 MacBook Pro with 32 GB of unified memory using the MLX framework by \cite{mlx:2023}.

\section{Results}
This section details our experimental findings. We first optimize the synthetic data generation parameters in \ref{results-data-parameters}. We then select the best LLM to create the dataset in \ref{results-generation-llm} and explore some of the created dataset statistics in \ref{results-optimal-dataset}. Finally, the most important results are presented in \ref{results-evaluation}, where we benchmark several models fine-tuned on this data. These show the final benchmark performance. Our results confirm that fine-tuning with our synthetic data dramatically improves model performance.
\subsection{Optimization of Data Generation Parameters} \label{results-data-parameters}
We first analyzed the impact of varying the number of synthetic fallacious premises ($K$) and synthetic claim/premise pairs ($M$) on the performance of a fine-tuned Phi-4 (8-bit) model by \cite{phi4:2024}. The results are presented in Table \ref{table:km-selection}.
\begin{table}[ht!]
\centering
\caption{Fine-tuned performance of Phi-4 (8-bit) with varying synthetic data parameters $K$ and $M$. LoRA layers are 16. All fine-tuning runs were executed for 500 iterations. Performance is measured on the \textit{MISSCI} test split.}
\label{table:km-selection}
\begin{tabular}{@{}cclccccc@{}}
\toprule
\thead{K} & \thead{M} & \thead{Generation LLM} & \thead{Val Loss \\ 1 iter} & \thead{Val Loss \\ 500 iters} & \thead{Acc} & \thead{F1 \\ (macro)} & \thead{Train \\ Samples} \\
\midrule
0  & 0  & Vanilla           & -     & -     & 0.667          & 0.550          & -    \\
10 & 0  & Random baseline   & 1.938 & 0.166 & 0.606          & 0.512          & 299  \\
10 & 0  & o4-mini           & 1.938 & 0.147 & 0.685          & 0.622          & 299  \\
15 & 5  & o4-mini           & 1.943 & 0.076 & 0.711          & 0.654          & 929  \\
30 & 15 & o4-mini           & 1.940 & 0.067 & \textbf{0.762} & \textbf{0.690} & 2344 \\
40 & 20 & o4-mini           & 1.943 & 0.074 & 0.711          & 0.647          & 2984 \\
\bottomrule
\end{tabular}
\end{table}

The vanilla Phi-4 model achieved an F1-score of 0.550. Fine-tuning consistently improved performance, demonstrating the effectiveness of the synthetic data, which is further validated by the significant drop in the validation loss from approximately 1.94 down to as low as 0.067 for all successful configurations. Notably, the \textit{Random baseline ablation} underperformed (F1 of 0.512), despite showing an initial Val Loss of 1.938 dropping to 0.166, confirming that the model learned from the synthetic data, rather than the prompt template or answer structure alone.

We observed a maximum F1-score of 0.690 and a maximum accuracy of 0.762 at $K=30$ and $M=15$. Increasing the data volume further to $K=40$ and $M=20$ led to a decrease in performance, with the F1-score dropping to 0.647. Considering this peak in performance and the associated generation/fine-tuning costs, we selected the configuration $K=30$ and $M=15$ for subsequent experiments. This configuration achieved a competitive F1-score of 0.690 (a 14\% absolute gain over the vanilla model), representing the optimal setting with 2344 training samples and a favorable validation loss reduction from 1.940 to 0.067.

\subsection{Selecting the Generation LLM}  \label{results-generation-llm}
Next, we investigated whether the quality of the LLM used for synthetic data generation impacts the final fine-tuned model's performance. Using the optimal parameters ($K=30, M=15$), we compared four different generator models (Table \ref{table:generation-llm-selection}).
\begin{table}[ht!]
\centering
\caption{Comparison of different LLMs used for generating synthetic data ($\mathcal{D}_{\mathrm{syn}}$) for Phi-4 (8-bit) fine-tuning. $K=30$, $M=15$, LoRA layers are 16. All fine-tuning runs were executed for 500 iterations.}
\label{table:generation-llm-selection}
\begin{tabular}{@{}llcccc@{}}
\toprule
\thead{Generation \\ LLM} & \thead{Fine-tuned \\ LLM} & \thead{Val Loss \\ 1 iter} & \thead{Val Loss \\ 500 iters} & \thead{Acc} & \thead{F1 \\ (macro)} \\
\midrule
o4-mini         & Phi-4 (8-bit) & 1.940 & 0.067 & 0.762 & 0.690 \\
GPT-4.1         & Phi-4 (8-bit) & 1.951 & 0.058 & 0.751 & 0.653 \\
GPT-5 (medium)  & Phi-4 (8-bit) & 1.945 & 0.063 & \textbf{0.764} & \textbf{0.705} \\
o3              & Phi-4 (8-bit) & 1.945 & 0.081 & 0.731 & 0.621 \\
\bottomrule
\end{tabular}
\end{table}

We observed that the data generated by GPT-5 (medium) resulted in the highest F1-score (0.705) and accuracy (0.764), demonstrating its superior ability to generate high-quality training examples. Therefore, prioritizing maximum performance for this task, we selected GPT-5 (medium) as the best generation model for \textit{MisSynth}, despite the higher generation cost.

\subsection{Optimal Synthetic Dataset} \label{results-optimal-dataset}
We publicly release our optimal synthetic dataset. We note that the dataset was generated once using the final iteration of the \textit{MisSynth} code with $K=30$ and $M=15$ (GPT-5). This dataset is used for all subsequent experiments. The dataset was generated once. Table \ref{table:fallacy-distribution} details the distribution of fallacy categories. The synthetic dataset's distribution differs from the \textit{MISSCI} splits. For instance, \textit{Fallacy of Exclusion} comprises a smaller portion (7.44\%) compared to the test split (27.53\%). Conversely, some minority classes in the \textit{MISSCI} test split, such as \textit{False Dilemma} (4.19\%) and \textit{Impossible Expectations} (1.32\%), are represented more frequently in the synthetic data (11.21\% and 8.44\%, respectively).

\begin{table}[ht!]
\centering
\caption{Distribution of Fallacy Categories (Count and Percentage) across datasets.}
\label{table:fallacy-distribution}
\begin{tabular}{@{}lccc@{}}
\toprule
\thead{Fallacy Category} & \thead{MISSCI, dev split} & \thead{MISSCI, test split} & \thead{GPT-5 $\mathcal{D}_{\mathrm{syn}}$ dataset \\ ($K=30, M=15$)} \\
\midrule
Ambiguity & 7 (7.29\%) & 44 (9.69\%) & 129 (14.32\%) \\
Biased Sample Fallacy & 10 (10.42\%) & 37 (8.15\%) & 84 (9.32\%) \\
Causal Oversimplification & 14 (14.58\%) & 73 (16.08\%) & 133 (14.76\%) \\
Fallacy of Division/Composition & 7 (7.29\%) & 33 (7.27\%) & 73 (8.10\%) \\
Fallacy of Exclusion & 25 (26.04\%) & 125 (27.53\%) & 67 (7.44\%) \\
False Dilemma & 8 (8.33\%) & 19 (4.19\%) & 101 (11.21\%) \\
False Equivalence & 14 (14.58\%) & 85 (18.72\%) & 115 (12.76\%) \\
Hasty Generalization & 6 (6.25\%) & 32 (7.05\%) & 123 (13.65\%) \\
Impossible Expectations & 5 (5.21\%) & 6 (1.32\%) & 76 (8.44\%) \\
\midrule
Overall & 96 (100.00\%) & 454 (100.00\%) & 901 (100.00\%) \\
\bottomrule
\end{tabular}
\end{table}
Table \ref{table:rouge} shows the ROUGE recall \citep{rouge:2004}, measuring textual overlap between entities and their source article excerpt $E_i$. The synthetic \textit{Context} (0.766) and \textit{Accurate Premise} (0.862) show higher recall than the \textit{MISSCI dev split} (0.635 and 0.741). In contrast, the synthetic \textit{Fallacy} (0.493) and \textit{Claim} (0.612) have lower recall than the dev split (0.608 and 0.642). This suggests that both synthetic data and \textit{MISSCI} is well-grounded in the source article. Textual comparison of dataset entities are available in Appendix \ref{appendix-c} tables \ref{appendix-c-table-1}, \ref{appendix-c-table-2}, \ref{appendix-c-table-3}.
\begin{table}[ht!]
\centering
\caption{ROUGE recall between article excerpt $E_i$ and entity.}
\label{table:rouge}
\begin{tabular}{@{}lccc@{}}
\toprule
\thead{Dataset Entity} & \thead{ROUGE \\ MISSCI dev split} & \thead{ROUGE \\ GPT-5 $\mathcal{D}_{\mathrm{syn}}$ dataset \\ ($K=30, M=15$)} \\
\midrule
Fallacy ($K$) & 0.608 & 0.493 \\
Context ($K$) & 0.635 & 0.766 \\
Claim ($M$) & 0.642 & 0.612 \\
Accurate Premise ($M$) & 0.741 & 0.862 \\
\bottomrule
\end{tabular}
\end{table}

\subsection{Evaluation of Fine-tuned Models} \label{results-evaluation}
Finally, we benchmarked the performance gains achieved by fine-tuning various LLMs using our GPT-5 $\mathcal{D}_{\mathrm{syn}}$ dataset ($K=30, M=15$). Table \ref{table:model-comparison} compares the vanilla and fine-tuned performance.

\begin{table}[ht!]
\centering
\caption{Comparison of different base models before and after fine-tuning with GPT-5 $\mathcal{D}_{\mathrm{syn}}$ ($K=30, M=15$). All fine-tuning runs were executed for 500 iterations. Performance measured on the \textit{MISSCI} test split.}
\label{table:model-comparison}
\begin{tabular}{@{}lcccccccc@{}}
\toprule
\thead{Fine-tuned LLM} & \thead{Val \\ Loss \\ 1 iter} & \thead{Val \\ Loss \\ 500 iters} & \thead{Vanilla \\ Acc} & \thead{Vanilla \\ F1 } & \thead{Fine \\ Acc} & \thead{Fine \\ F1 } & \thead{LLM \\ Size} & \thead{LoRA \\ Layers} \\
\midrule
Gemma 3 (8-bit) & 3.324 & 0.067 & 0.531 & 0.377 & 0.764 & 0.691  & 4B  & 32 \\
LLaMA 3.1 (4-bit) & 2.451 & 0.050 & 0.414 & 0.334 & \textbf{0.778} & 0.711  & 8B  & 32 \\
LLaMA 2 (4-bit) & 2.145 & 0.073 & 0.326 & 0.218 & 0.722 & 0.681  & 13B & 32 \\
Phi-4 (8-bit) & 1.945 & 0.063 & 0.667 & 0.550 & 0.764 & 0.705  & 15B & 16 \\
Mistral Small 3.2 (4-bit) & 2.124 & 0.072 & 0.698 & 0.553 & 0.762 & \textbf{0.718}  & 24B & 16 \\
LLaMA 2 *  & -     & -     & 0.577 & 0.464 & -     & -      & 70B & - \\
GPT-4 *   & -     & -     & \textbf{0.738} & \textbf{0.649} & -     & -      & -   & - \\
\bottomrule
\end{tabular}
\footnotetext{* Results by \cite{missci:2024}}
\end{table}

The results confirm that the \textit{MisSynth} significantly improves performance across different model architectures. All fine-tuned models showed substantial decreases in validation loss, with LLaMA 3.1 (8B) by \cite{llama3:2024} dropping from 2.451 to 0.050 and Gemma 3 (4B) by \cite{gemma3:2025} dropping from 3.324 to 0.067, indicating successful adaptation to the fallacy classification task.

The LLaMA 2 13B model \citep{llama:2023} showed the largest absolute improvement, increasing its F1-score from a baseline of 0.218 to 0.681, alongside a validation loss reduction from 2.145 to 0.073. The fine-tuned Mistral Small 3.2 model achieved the highest F1-score overall at 0.718 (a 16.5\% absolute gain). Other models also showed strong performance, with LLaMA 3.1 achieving 0.711 (37.7\% absolute gain) and Phi-4 reaching 0.705 (15.5\% absolute gain). Notably, several fine-tuned smaller models outperformed the proprietary model. Our fine-tuned Mistral Small 3.2 \citep{mistral:2024}, LLaMA 3.1, Phi-4, and Gemma 3 (F1 of 0.691) all surpassed the vanilla GPT-4 model \citep{gpt4:2024}, which was reported to have an F1 of 0.649.

Due to VRAM limitations, we were unable to fine-tune or evaluate the LLaMA 2 70B \citep{llama:2023} model directly. Therefore, the reported vanilla performance for the LLaMA 2 70B (F1: 0.464, Acc: 0.577) and GPT-4 (F1: 0.649, Acc: 0.738) is taken from Table 3 of the original MISSCI paper \citep{missci:2024}. Critically, the fine-tuned LLaMA 2 13B (F1: 0.681) substantially outperformed the vanilla, much larger LLaMA 2 70B model (F1: 0.464). This highlights a key finding: targeted training using high-quality, RAG-supported synthetic data can close the performance gap between small, parameter-efficient models and large foundation models for domain-specific tasks like fallacy classification.

\begin{table}[ht!]
\centering
\caption{Comparison of Vanilla vs. Fine-Tuned LLaMA 2 13B F1-Scores by Fallacy Category on the \textit{MISSCI} test split.}
\label{table:model-category-f1}
\begin{tabular}{@{}lcccc@{}}
\toprule
\thead{Fallacy Category} & \thead{Count} & \thead{Vanilla \\ F1 (macro)} & \thead{Fine-Tuned \\ F1 (macro)} & \thead{Absolute \\ Gain} \\
\midrule
Ambiguity & 44 & 0.044 & 0.333 & 0.289 \\
Biased Sample Fallacy & 37 & 0.143 & 0.704 & 0.561 \\
Causal Oversimplification & 73 & 0.485 & 0.820 & 0.335 \\
Fallacy of Division/Composition & 33 & 0.050 & 0.485 & 0.435 \\
Fallacy of Exclusion & 125 & 0.110 & 0.954 & \textbf{0.844} \\
False Dilemma & 19 & 0.148 & 0.812 & 0.664 \\
False Equivalence & 85 & 0.614 & 0.479 & -0.135 \\
Hasty Generalization & 32 & 0.586 & 0.912 & 0.326 \\
Impossible Expectations & 6 & 0.000 & 0.632 & 0.632 \\
\midrule
Macro Average F1 & 454 & 0.218 & 0.681 & 0.463 \\
Accuracy & 454 & 0.326 & 0.722 & 0.396 \\
\bottomrule
\end{tabular}
\end{table}

The category-specific analysis of LLaMA 2 13B model results reveals that the largest absolute improvement in macro F1-score from 0.218 to 0.681 across all fallacy categories is driven by dramatic performance improvements across nearly all categories, especially those where the vanilla model was weakest. The most significant improvements were seen in \textit{Fallacy of Exclusion}, which rose from an F1-score of 0.110 to 0.954, and \textit{False Dilemma}, which increased from 0.148 to 0.812. Furthermore, the model learned to identify the highly minority class \textit{Impossible Expectations}, improving from an F1-score of zero to 0.632. Strong gains were also observed in other low-performing categories such as \textit{Biased Sample Fallacy} (0.143 to 0.704). Notably, performance on \textit{False Equivalence} decreased from 0.614 to 0.479 after fine-tuning. Overall, the results demonstrate that our synthetic data generation pipeline is highly effective at strengthening model performance, particularly on challenging fallacy classes, significantly improving the model's overall robustness, F1 score and accuracy.

\section{Discussion}
We introduced \textit{MisSynth}, a novel pipeline for generating high-quality synthetic data to detect scientific fallacies. Our method significantly improves the performance of LLMs on the \textit{MISSCI} benchmark. Fine-tuning even small models, such as LLaMA 3.1 8B, with our data yielded substantial gains, surpassing the performance of much larger vanilla models, like GPT-4. This demonstrates that targeted, parameter-efficient fine-tuning with context-aware synthetic data is an effective strategy for specialized reasoning tasks.
\subsection{Limitations}
Our research's primary limitation is its exclusive focus on the \textit{MISSCI} benchmark by \cite{missci:2024}. Consequently, our synthetic data and fine-tuned models are specialized for this dataset. Furthermore, our methodology addresses only the classification sub-task. We do not evaluate the generation of fallacious premises, which is another part of the \textit{MISSCI} dataset.
\subsection{Future work}
Future work includes generalizing \textit{MisSynth}. We aim to adapt the method for other fallacy benchmarks, such as those mentioned in our related work, like \textit{MAFALDA} by \cite{mafalda:2024}. We also plan to scale our solution. This involves moving beyond local hardware to fine-tune larger models on cloud infrastructure.
\subsection{Ethical considerations}
Our synthetic dataset was generated automatically by an LLM. No medical experts or health professionals reviewed the synthetic data. There is a potential danger that malicious actors could exploit our synthetic data to spread health misinformation more effectively.

\backmatter
\bmhead{Acknowledgements}

We thank Max Glockner for validating the initial idea and appreciate his helpful feedback during the development of this work.

\begin{appendices}
\section{Single Class Synthetic Fallacy Prompt Template} \label{appendix-a}
You are provided with a claim, an accurate premise for the claim, a list of real-world fallacious premises (fallacies)
from the scientific article with the fallacy class, and relevant text exempt from this article.
\par\medskip
\noindent Claim: \texttt{\{claim\}}
\noindent Accurate Premise: \texttt{\{premise\}}
\par\medskip
\noindent \texttt{\{fallacies\}}
\par\medskip
\noindent Article Excerpt: \texttt{\{article\_excerpt\}}
\par\medskip
\noindent Task:
\par\medskip
\noindent Based on the example and relevant text from the article, create \texttt{\{n\_entries\}} synthetic fallacies that
differ from the provided real-world fallacies and their class in the JSON format:
\begin{verbatim}
[
    {
        "context": // Synthetic Context 1,
        "fallacy": // Synthetic Fallacy 1,
        "class": // Synthetic Class 1
    },
    {
        "context": // Synthetic Context 2,
        "fallacy": // Synthetic Fallacy 2,
        "class": // Synthetic Class 2
    },
    ...
    {
        "context": // Synthetic Context {n_entries},
        "fallacy": // Synthetic Fallacy {n_entries},
        "class": // Synthetic Class {n_entries}
    }
]
\end{verbatim}

\noindent Creating fallacies of the classes different from provided real-world examples is encouraged,
but the class could be only from the fallacy inventory.
\par\medskip
\noindent \texttt{\{fallacy\_inventory\}}
\par\medskip
\noindent Structure created fallacy text similarly to real-world examples.
\section{Synthetic Claim-Accurate Premise Prompt Template}\label{appendix-b}
You are provided with a claim, an accurate premise, a list of real-world fallacious premises (fallacies)
from the scientific article with the fallacy class, and relevant text exempt from this article.
\par\medskip
\noindent Claim: \texttt{\{claim\}}
\noindent Accurate Premise: \texttt{\{premise\}}
\par\medskip
\noindent \texttt{\{fallacies\}}
\par\medskip
\noindent Article Excerpt: \texttt{\{article\_excerpt\}}
\par\medskip
\noindent Task:
\par\medskip
\noindent Based on the example and relevant text from the article, create \texttt{\{n\_entries\}} synthetic claim and accurate premise pairs that
differ from the provided real-world premises in the JSON format. Make sure that the created claim-accurate premise pair is coherent.
\begin{verbatim}
[
    {
        "premise": // Synthetic Accurate Premise 1,
        "claim": // Synthetic Claim 1,
    },
    {
        "premise": // Synthetic Accurate Premise 2,
        "claim": // Synthetic Claim 2,
    },
    ...
    {
        "premise": // Synthetic Accurate Premise {n_entries},
        "claim": // Synthetic Claim {n_entries}
    }
]
\end{verbatim}

\noindent Structure created claims and accurate premises text similarly to real-world examples.

\section{Optimal Synthetic Dataset Examples} \label{appendix-c}
\begin{table}[ht!]
\centering
\caption{Comparison of \textit{MISSCI} dev split vs. randomly chosen Synthetic Claim-Accurate Premise pairs (GPT-5 $\mathcal{D}_{\mathrm{syn}}$ dataset $K=30, M=15$, argument ID 171)}
\label{appendix-c-table-1}
\begin{tabularx}{\textwidth}{@{}lXXXX@{}}
\toprule
\thead{Source} & \thead{Claim} & \thead{Accurate \\ Premise} \\
\midrule
\textit{MISSCI} dev split & COVID-19 immunity likely lasts for years. & Different types of immune cells contributing to immune memory and long-term protection remained detectable in the blood of recovered COVID-19 patients \\

Synthetic & SARS-CoV-2 T cell memory may stabilize rather than rapidly decline over time. & Data suggest T cell memory may reach a stable plateau beyond the first eight months after infection. \\

Synthetic & Long-term protection from COVID-19 depends on durable immune memory. & Immune memory is the source of long-term protective immunity against reinfection. \\

Synthetic & Definitive conclusions about the duration of COVID-19 immunity are still premature. & The overall amount of data on protective immunity to SARS-CoV-2 remains limited. \\

\bottomrule
\end{tabularx}
\end{table}

\begin{sidewaystable}[ht!]
\centering
\caption{Comparison of \textit{MISSCI} dev split vs. randomly chosen Synthetic Claim-Accurate Premise pairs (GPT-5 $\mathcal{D}_{\mathrm{syn}}$ dataset $K=30, M=15$, argument ID 171). "N/A" means that the entity is not present in the argument. Part 1}
\label{appendix-c-table-2}
\begin{tabularx}{\textheight}{@{}XXXXX@{}}
\toprule
\thead{Fallacy Category} & \thead{\textit{MISSCI} dev split \\ Fallacy Example} & \thead{\textit{MISSCI} dev split \\ Context Example} & \thead{Synthetic Fallacy \\ Example} & \thead{Synthetic Context \\ Example} \\
\midrule
Ambiguity & Stating that COVID-19 immunity likely lasts "for years" is precise enough to understand that antibodies, memory B cells, and memory T cells were found five to eight months after infection. & Different types of immune cells remained detectable in the blood of recovered COVID-19 patients for up to eight months. & Blunting disease severity is the same as having full immunity for years. & Sub-sterilizing neutralizing antibody titers can blunt the size of the initial infection and limit COVID-19 severity. \\

Biased Sample Fallacy & N/A & N/A & Citing extreme cases of long-lived memory implies typical COVID-19 immunity is similarly long-lived. & Reports of extremely long-lived B cell memory to other infections (e.g., smallpox, influenza). \\

Causal Oversimplification & Antibodies, memory B cells, and memory T cells provide the main protection against viral infections. Therefore, because they were found months after infection, COVID-19 immunity likely lasts for years. & N/A & A few months of stability cause a guaranteed linear extension into multiple years. & RBD-specific memory B cells showed no apparent decline over 5-8 months. \\

Fallacy of Division/Composition & Antibodies, memory B cells and memory T cells are part of the immune system. Therefore, the immune system lasts for years if antibodies last for years. & N/A & Because memory cells are present in blood, every tissue and mucosal site will be protected for years. & Detection of immune memory cells in blood up to 8 months post-infection. \\
\bottomrule
\end{tabularx}
\end{sidewaystable}
\begin{sidewaystable}[ht!]
\centering
\caption{Comparison of Real-World vs. randomly chosen Synthetic Fallacy Examples by Category (GPT-5 $\mathcal{D}_{\mathrm{syn}}$ dataset $K=30, M=15$, argument ID 171). "N/A" means that the entity is not present in the argument. Part 2}
\label{appendix-c-table-3}
\begin{tabularx}{\textheight}{@{}XXXXX@{}}
\toprule
\thead{Fallacy Category} & \thead{\textit{MISSCI} dev split \\ Fallacy Example} & \thead{\textit{MISSCI} dev split \\ Context Example} & \thead{Synthetic Fallacy \\ Example} & \thead{Synthetic Context \\ Example} \\
\midrule
Fallacy of Exclusion & It is irrelevant to the claim that different types of immune cells remained detectable in the blood of recovered COVID-19 patients for up to eight months.& Different types of immune cells remained detectable in the blood of recovered COVID-19 patients for up to eight months. & By focusing only on cohorts with robust responses and short follow-up, we can conclude multi-year immunity. & Several cohorts detected robust RBD memory B cells within the first 8 months. \\
False Dilemma & Either something vanishes quickly or stays for years. & Different types of immune cells remained detectable in the blood of recovered COVID-19 patients for up to eight months & Either immunity blocks all infection or it is worthless; since memory cells are present, they must block infection for years. & Minimizing COVID-19 disease severity by confining SARS-CoV-2 to the upper respiratory tract is a primary goal mediated by memory T and B cells. \\

False Equivalence & N/A & N/A & Since some pathogens induce decades-long B cell memory, SARS-CoV-2 immunity will last decades too. & B cell memory to smallpox vaccination can last 60+ years, and after influenza infection 90+ years. \\

Hasty Generalization & N/A & N/A & Therefore, SARS-CoV-2 immunity lasts at least 17 years in humans. & SARS-CoV T cells have been detected 17 years after the initial infection. \\

Impossible Expectations & N/A & N/A Because it is impossible to have 10-year follow-up data right now, we should accept that immunity lasts for years based on months of data. & Conclusions are constrained by the limited overall amount of data on protective immunity to SARS-CoV-2. \\
\bottomrule
\end{tabularx}
\end{sidewaystable}
\end{appendices}

\bibliography{sn-bibliography}
\end{document}